\documentclass[dvipsnames]{article} %
\usepackage{colm2024_conference}

\usepackage{booktabs}
\usepackage{graphicx}
\usepackage{enumitem}
\usepackage{wrapfig}
\usepackage{algorithm}
\usepackage{algpseudocode}

\usepackage{microtype}
\usepackage{amsmath}
\usepackage{colortbl}
\usepackage[utf8]{inputenc}
\definecolor{lightgray}{rgb}{0.9,0.9,0.9}
\usepackage{caption}
\usepackage{subcaption}
\usepackage{setspace}
\usepackage{url}
\usepackage{multirow}
\usepackage{colortbl}
\usepackage{tabularx}
\usepackage{blindtext}
\usepackage{pgfplots}
\pgfplotsset{compat=1.18} 
\usepackage{tikz}
\usetikzlibrary{er,positioning,bayesnet}
\usepackage{makecell}
\usepackage{tipa}
\usepackage{siunitx}
\usepackage{nicefrac}
\usepackage{tocloft}
\usepackage{listings}
\usepackage[raster,skins]{tcolorbox} %
\usepackage{xltabular}
\usepackage{adjustbox}
\usepackage{xurl}
\usepackage{rotating}
\usepackage[normalem]{ulem}
\useunder{\uline}{\ul}{}

\usepackage{amsmath,amsfonts,bm}

\def\eqref#1{equation~\ref{#1}}
\def\1{\bm{1}}

\DeclareMathAlphabet{\mathsfit}{\encodingdefault}{\sfdefault}{m}{sl}
\SetMathAlphabet{\mathsfit}{bold}{\encodingdefault}{\sfdefault}{bx}{n}

\newcommand*\justify{%
  \fontdimen2\font=0.4em% interword space
  \fontdimen3\font=0.2em% interword stretch
  \fontdimen4\font=0.1em% interword shrink
  \fontdimen7\font=0.1em% extra space
  \hyphenchar\font=`\-% allowing hyphenation
}

\renewcommand{\texttt}[1]{%
  \begingroup
  \ttfamily
  \begingroup\lccode`~=`/\lowercase{\endgroup\def~}{/\discretionary{}{}{}}%
  \begingroup\lccode`~=`[\lowercase{\endgroup\def~}{[\discretionary{}{}{}}%
  \begingroup\lccode`~=`.\lowercase{\endgroup\def~}{.\discretionary{}{}{}}%
  \catcode`/=\active\catcode`[=\active\catcode`.=\active
  \justify\scantokens{#1\noexpand}%
  \endgroup
}

\title{Egocentric Gaze Estimation via Neck-Mounted Camera}

\author{%
\makebox[0.45\linewidth][c]{\textbf{Haoyu Huang}}%
\makebox[0.45\linewidth][c]{\textbf{Yoichi Sato}}\\[0.25em]
\makebox[0.45\linewidth][c]{The University of Tokyo}%
\makebox[0.45\linewidth][c]{The University of Tokyo}\\
\makebox[0.45\linewidth][c]{\texttt{hyhuang@iis.u-tokyo.ac.jp}}%
\makebox[0.45\linewidth][c]{\texttt{ysato@iis.u-tokyo.ac.jp}}%
}

\begin{document}

\maketitle

\begin{abstract}
This paper introduces neck-mounted view gaze estimation, a new task that estimates user gaze from the neck-mounted camera perspective. Prior work on egocentric gaze estimation, which predicts device wearer’s gaze location within the camera’s field of view, mainly focuses on head-mounted cameras while alternative viewpoints remain underexplored. To bridge this gap, we collect the first dataset for this task, consisting of approximately 4 hours of video collected from 8 participants during everyday activities. We evaluate a transformer-based gaze estimation model, GLC, on the new dataset and propose two extensions: an auxiliary gaze out-of-bound classification task and a multi-view co-learning approach that jointly trains head-view and neck-view models using a geometry-aware auxiliary loss. Experimental results show that incorporating gaze out-of-bound classification improves performance over standard fine-tuning, while the co-learning approach does not yield gains. We further analyze these results and discuss implications for neck-mounted gaze estimation.

\end{abstract}

% \vfill

% \newpage

\section{Introduction}

Egocentric gaze estimation is one of the sub-tasks in the field of egocentric (first-person view) vision and human behavior understanding. 
Given a first-person perspective video, the model is trained to predict where the user's gaze is located inside the image frame for each timestep.
Gaze locations provide strong hints for human attention and can be utilized in various downstream tasks such as action recognition or object segmentation \citep{li2013learning}.

Prior works and major datasets in egocentric gaze estimation usually utilize video sources that are recorded from head-mounted cameras. 
With the progress of the egocentric recording devices, however, non-head mounted variants such as neck-mounted cameras have emerged and being utilized in egocentric vision tasks \citep{thinklet2025}. 
Neck-mounted cameras stand out in terms of comfort and offer improved usability in certain specialized working conditions compared with head-mounted devices. 
For example, necklace-like neck-mounted cameras have been utilized by an air-conditioning company to capture egocentric videos from their field technicians \citep{daikin2024thinklet}.
However, despite the unique advantages of neck-mounted cameras, to the best of our knowledge, no existing work has explored the problem of gaze estimation using footage captured from neck-mounted cameras.

Driven by the motivation of bridging the gap, we propose the new task of \textbf{neck-mounted view gaze estimation}, which is defined as predicting the user's gaze location from egocentric footage recorded from a neck-mounted wearable device.
To provide a starting point for the proposed task, we recorded the first dataset for neck-mounted view gaze estimation.
We used a co-collection approach by combining an eye-tracker equipped head-mounted camera Aria \citep{engel2023aria} and a monocular neck-mounted camera THINKLET \citep{thinklet2025}. 
Those two cameras are time-synchronized to collect paired RGB video footage and gaze data inside the head-mounted camera's coordinate frame.
We obtain the gaze point in the neck-mounted camera's coordinate frame by feeding the paired video frame and gaze into VGGT \citep{wang2025vggt}, which is a 3D foundation model that is capable of inferring point correspondence between different camera views.
Our dataset contains 4 hours of effective video footage from 8 participants, recording daily activities such as preparing coffee or Jenga gameplays, and annotated with gaze coordinates, gaze-in-bound classification labels and estimated relative translations.

We used the collected dataset to evaluate existing models' performance on neck-mounted gaze estimation. Specifically, we report performance by directly fine-tuning the Global-local correlation model (GLC) \citep{lai2024glc}, a transformer-based model originally designed for head-mounted gaze estimation.
Additionally, we propose two domain-specific augmentation methodologies.
The first approach is incorporating an auxiliary gaze-in-bound classification task, 
i.e. classifying whether the user's gaze is inside the camera field of view, 
in addition to the original heatmap-based gaze localization task.
This is driven by the preliminary analysis on the dataset which shows a substantially higher out-of-bound rate comparing to head-mounted view setup. 
We hypothesize that the auxiliary task provides extra supervision signal for the main gaze estimation model.
The second approach is a multi-view co-learning approach that leverages synchronized head-mounted footage as extra supervision, where two models with shared architecture are jointly trained on head-view and neck-view data, respectively. We introduce an auxiliary loss that aligns the models' bottleneck representations in addition to the normal heatmap loss. Following prior latent feature alignment methods \citep{spurr2021peclr,hisadome2024rotation}, we project bottleneck features into a 3D-like latent space and compute the alignment loss conditioned the extrinsic rotation matrix between two cameras.

Experiments on the collected dataset shows that the auxiliary classification approach brings modest improvement comparing to the vanilla finetuning baseline, while co-learning approach does not yield gains. 
We also provide qualitative visualizations and discuss directions for future work.

To summarize, our contributions are as follows:
\begin{itemize}
  \item We introduce the task of neck-mounted view gaze estimation.
  \item We collect and analyze the first dataset for the new task and benchmark a modern gaze estimation model on neck-mounted views.
  \item We explore two domain-specific augmentation strategies for neck-mounted gaze estimation, namely auxiliary classification and multi-view co-learning, and provide insights into their effectiveness.
\end{itemize}

% ====== %
\section{Related Work}

\textbf{Egocentric vision.}
Egocentric (first-person view) vision studies visual data captured from wearable cameras that reflect the user’s perspective and interaction with the environment \citep{li2025egocentricsurvey}. This unique viewpoint enables the modeling of human activities, interactions, and attention, and has been widely explored in tasks such as action recognition \citep{li2018eye,min2021integrating,sudhakaran2018attention}, hand–object interaction \citep{liu2022hoi4d}, pose estimation \citep{mueller2017real,ohkawa2023assemblyhands}, and social interaction understanding \citep{northcutt2020egocom}. To support these efforts, a variety of egocentric datasets have been introduced, ranging from early datasets \citep{pirsiavash2012detecting} to large-scale multimodal benchmarks that incorporate gaze and other sensory signals \citep{grauman2022ego4d,ma2024nymeria}. Among these tasks, gaze estimation plays a central role as a proxy for user attention.

\textbf{Egocentric gaze estimation.}
Egocentric gaze estimation aims to predict the camera wearer’s gaze location within the image frame using only first-person video, without relying on explicit eye-tracking or eye appearance cues. 
One straightforward approach is leveraging visual saliency, motivated by the close relationship between gaze and visual attention.
However, studies \citep{yamada2011attention} have shown that saliency alone is insufficient, as gaze behavior is strongly influenced by user motion and task context. 
To address this, later work incorporated ego-motion cues, action cues and temporal dynamics \citep{yamada2011attention,fathi2012learning,huang2018predicting}, demonstrating improved performance by modeling user behavior more holistically. 
More recent methods adopt end-to-end methods with modern architecture like transformer-based models, achieving state-of-the-art level results on head-mounted gaze datasets \citep{lai2024glc}. 
Despite this progress, existing gaze estimation research overwhelmingly assumes head-mounted cameras.

\textbf{Non-head-mounted egocentric cameras.}
While head-mounted cameras provide strong alignment between camera motion and gaze, alternative camera placements such as chest- or neck-mounted devices have gained attention due to their improved comfort and practicality \citep{yamazoe2005body,yamazoe2007body}. Chest-mounted cameras have been used in egocentric activity understanding \citep{pirsiavash2012detecting,tadesse2022bon} and head pose estimation \citep{yamada2024head}, demonstrating their feasibility. 
More recently, neck-mounted cameras and other non-head-mounted perspectives have been adopted in real-world applications and egocentric vision studies \citep{ohnishi2016recognizing,thinklet2025,daikin2024thinklet}. 
However, to the best of our knowledge, gaze estimation from neck-mounted camera views has not been previously explored. Our work addresses this gap by studying neck-mounted view gaze estimation and analyzing its unique challenges.

\section{Neck-mounted View Gaze Estimation}

\subsection{Task Definition}
Neck-mounted view gaze estimation aims to predict the device wearer’s gaze location from egocentric video captured by a neck-mounted camera. Given a video clip $\mathcal{V}=\{V_t\}_{t=1}^{T}$, where $V_t \in \mathbb{R}^{H \times W \times 3}$ denotes the RGB frame at timestep $t$, the model outputs a sequence of gaze probability maps $\mathcal{P}=\{P_t\}_{t=1}^{T}$, with $P_t \in \mathbb{R}^{H \times W}$. Each heatmap encodes the likelihood of the gaze being at each pixel location. 

\subsection{Challenges}
Compared to head-mounted gaze estimation, neck-mounted views present several unique challenges. 
\begin{itemize}
    \item \textbf{Absence of public datasets.} To the best of our knowledge, there is no publicly available dataset for neck-mounted gaze estimation.
    \item \textbf{Different motion patterns.} Neck-mounted view cameras capture different motion patterns comparing to head-mounted view cameras. While head-mounted view camera's captured motion is strongly correlated with the wearer's head pose transition, neck-mounted view captures subtler movements. This brings additional challenge to the gaze estimation task.
    \item \textbf{Reduced center bias and frequent out-of-bound gazes.} The center bias \citep{li2013learning} commonly found in head-mounted gaze datasets is weaker in the neck-mounted setting, meaning gaze locations are less likely to concentrate near the center of the camera’s field of view. In addition, gaze points frequently fall outside the camera field of view in neck-mounted view camera setup. This out-of-bound phenomenon is rare in head-mounted settings but common in neck-mounted views, and poses additional challenges for heatmap-based supervision.
\end{itemize}

\subsection{Proposed Methods}
To adapt gaze estimation models pretrained on head-mounted data to the neck-mounted domain, we propose two domain-specific augmentation strategies. Both methods are model-agnostic and applicable to architectures following an encoder–bottleneck–decoder design.

\subsubsection{Auxiliary In-view Classification}
\begin{figure}[h]
    \centering
    \includegraphics[width=0.9\linewidth]{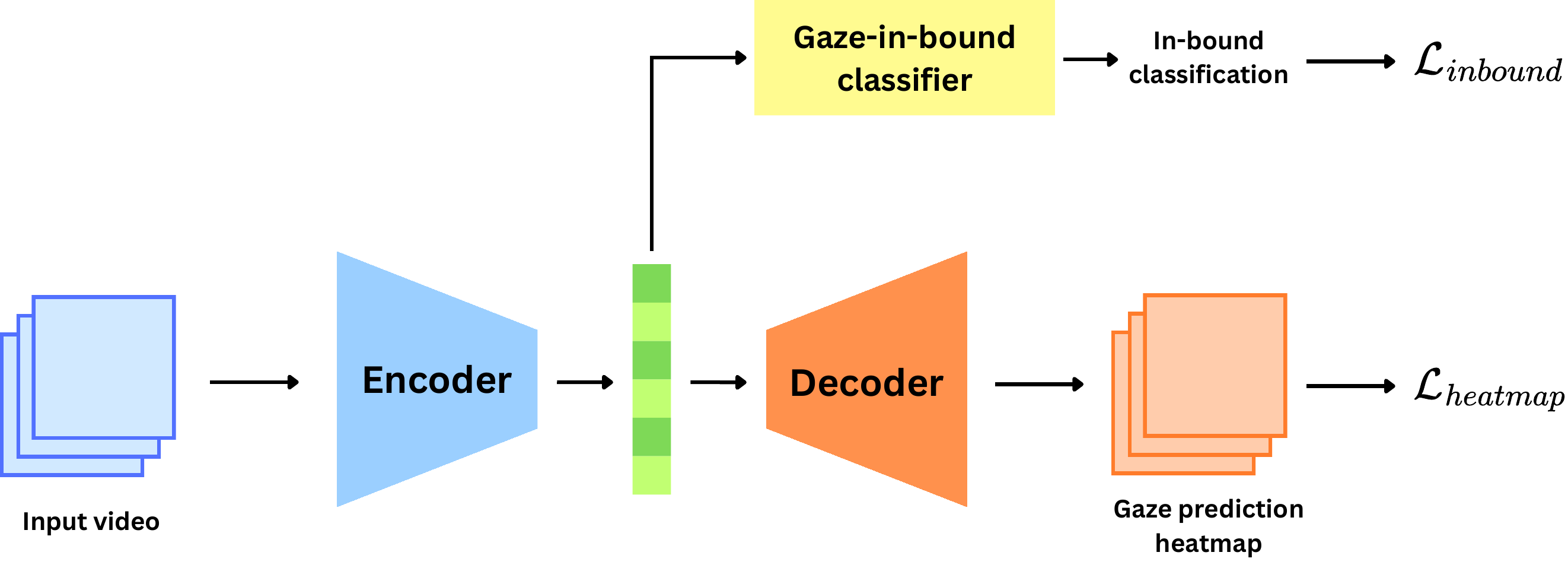}
    \caption{Architecture for the Auxiliary In-view Classification method. The Gaze-in-bound classifier branches out from the bottleneck feature of the original model and performs a binary classification task which classifies whether the gaze is inside the camera field-of-view. The in-bound classification loss is then used in backpropagation together with the heatmap loss.}
    \label{fig:aux_classif_pipeline}
\end{figure}

Existing gaze estimation models typically treat out-of-bound gaze as a special case, often supervising such frames with uniform heatmaps. In neck-mounted views, however, out-of-bound gaze occurs frequently.
To provide a clearer supervision signal, we introduce an auxiliary \emph{in-view classification} task as shown in Fig. \ref{fig:aux_classif_pipeline}. In addition to predicting gaze heatmaps $\mathcal{P}$, the model outputs a sequence of confidence scores $\mathcal{C}=\{C_t\}_{t=1}^{T}$, where $C_t$ denotes the probability that the gaze lies within the camera’s field of view at timestep $t$. The model is trained jointly using a heatmap regression loss $\mathcal{L}_{\text{heatmap}}$ and a binary classification loss $\mathcal{L}_{\text{inbound}}$. 
This auxiliary task encourages the model to learn the concept of out-of-bound gaze and we expect the awareness of gaze visibility will improve the heatmap localization performance.

\subsubsection{Multi-view Co-learning}
\begin{figure}[h]
    \centering
    \includegraphics[width=0.9\linewidth]{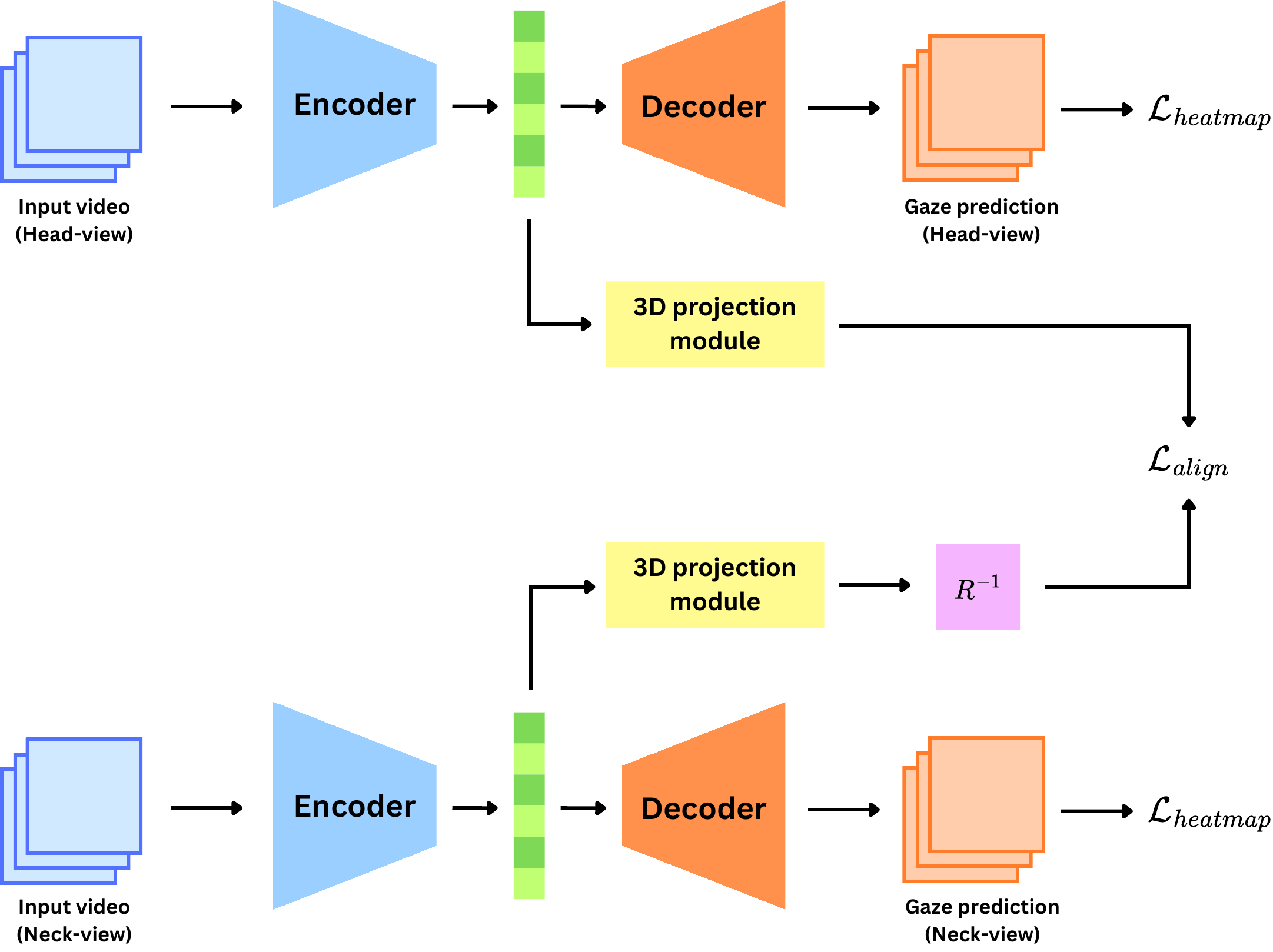}
    \caption{Architecture for the Multi-view Co-learning method. Two GLC \citep{lai2024glc} models are given with footage and gaze ground truth from head-view camera and neck-view camera, correspondingly. The 3D projection module projects the bottleneck feature into a 3D latent vector field. The vector field is then rotated by the relative extrinsic rotation matrix and an alignment loss is calculated between the rotated features. The alignment loss is then used together with the heatmap loss in the backpropagation.}
    \label{fig:multiview_pipeline}
\end{figure}

Given that most pretrained gaze models rely on head-mounted data, we further explore leveraging synchronized head-mounted footage as additional supervision. Using paired head- and neck-mounted video clips, we jointly train two models with shared architecture but separate parameters.
The designed architecture is illustrated in Fig. \ref{fig:multiview_pipeline}
Beyond the standard heatmap loss applied to each stream, we introduce an auxiliary alignment loss that encourages consistency between the bottleneck representations of the head-view and neck-view models.
Inspired by prior latent-space alignment methods \citep{spurr2021peclr,hisadome2024rotation}, we designed the alignment loss to condition on the relative extrinsic rotation between two camera's view. 
Specifically, we project bottleneck features into a 3D-like latent space and apply the relative rotation derived from the extrinsic calibration between the two cameras. The alignment loss is then computed between the rotated features.

\section{Dataset}

We collect a neck-mounted gaze dataset using a co-collection setup with an eye-tracker–equipped head-mounted camera and a monocular neck-mounted camera. Since neck-mounted devices cannot directly record gaze, we adopt the approach of inferring neck-view gaze by mapping head-view gaze annotations across synchronized camera views.

\subsection{Data Collection}
\begin{figure}[h]
    \centering
    \includegraphics[width=0.9\linewidth]{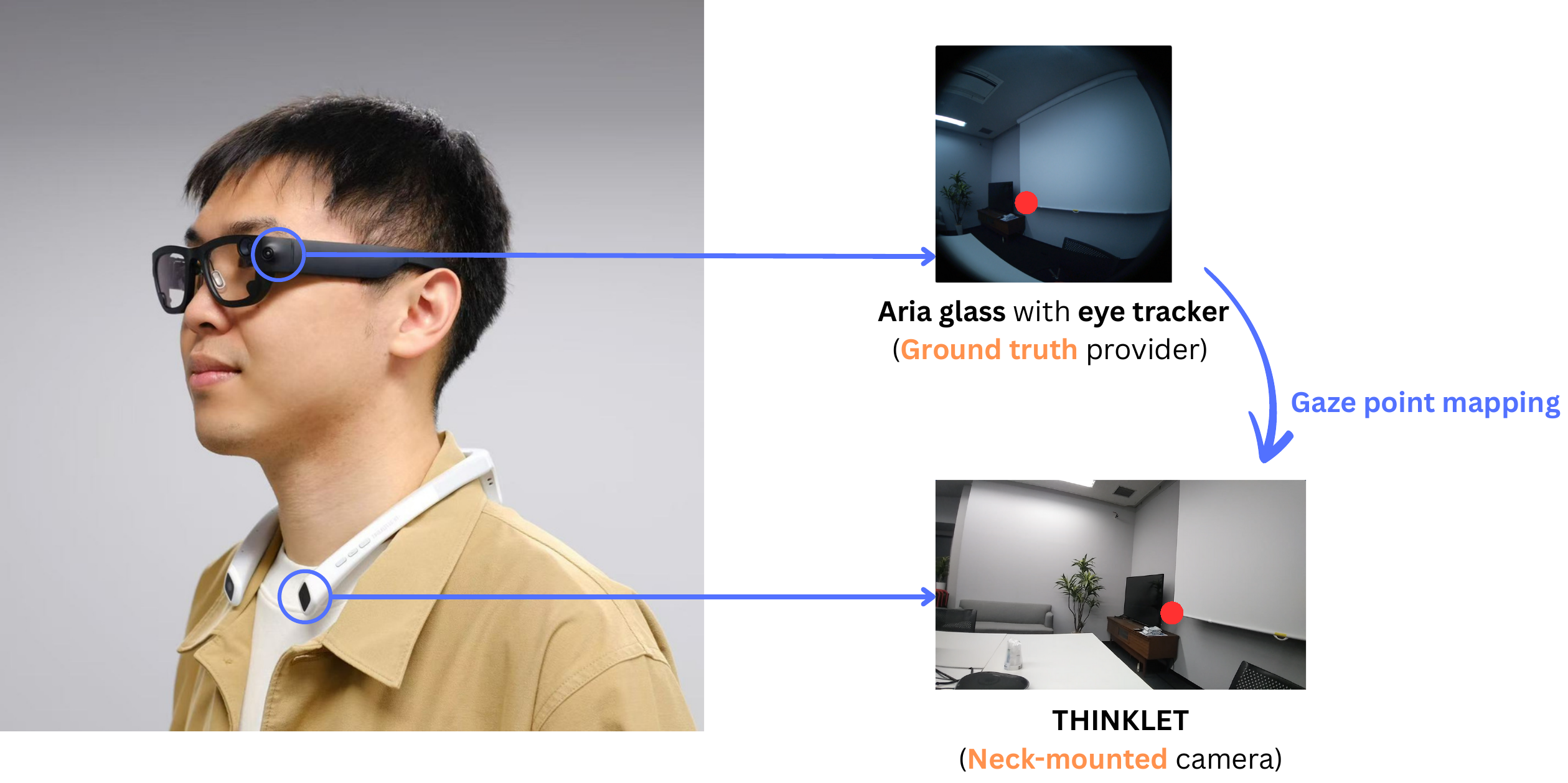}
    \caption{Collection methodology for the neck-mounted view gaze estimation dataset. The user wears an eye-tracker-equipped head-mounted camera and a neck-mounted camera, which records video simultaneously during the collection. The gaze point in the head-view camera coordinate is then mapped into the neck-mounted view camera coordinate frame.}
    \label{fig:collection_setup}
\end{figure}

The dataset is collected using Aria \citep{engel2023aria} as the head-mounted eye-tracking and recording device and Thinklet \citep{thinklet2025} as the neck-mounted recording device. The example setup is illustrated in Fig. \ref{fig:collection_setup}. Both devices record video simultaneously during the collection session. A total of 8 participants participated in the data collection, where they perform daily activities, including individual activities like preparing food and social interaction activities like Jenga gameplay. The list of tasks is shown in Table \ref{tab:task_distribution_time}.

\subsection{Gaze Annotation}
\begin{figure}[h]
    \centering
    \includegraphics[width=0.95\linewidth]{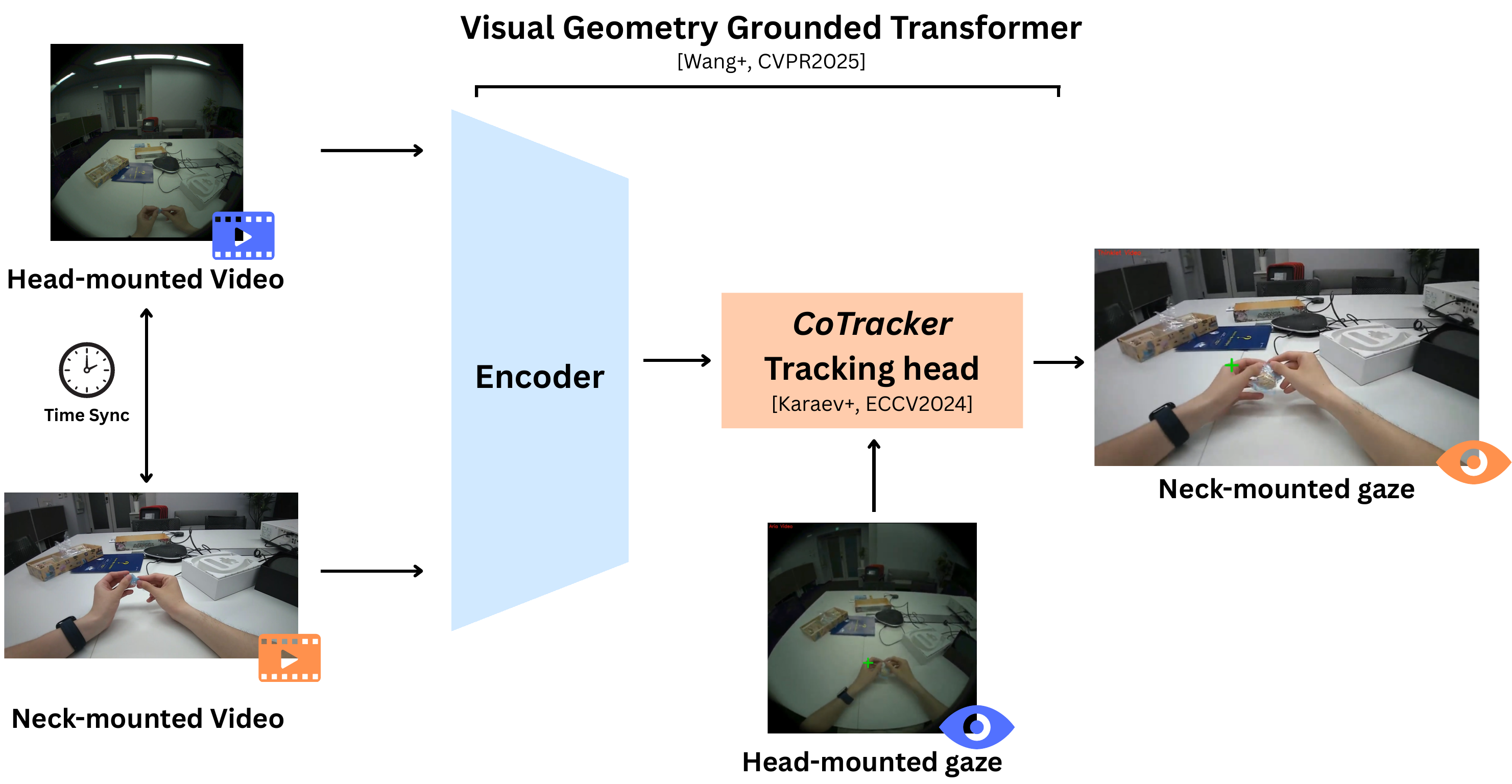}
    \caption{The gaze annotation pipeline for our dataset. For each timestep, a paired frame from synchronized head-view and neck-view videos are fed into the VGGT \citep{wang2025vggt} encoder. The CoTracker \citep{karaev2024cotracker} tracking head then takes the feature and the gaze point in the head-mounted view coordinate frame, and outputs the gaze point in the neck-mounted view coordinate frame.}
    \label{fig:mapping_pipeline}
\end{figure}

The overview of the gaze annotation pipeline is shown in Fig. \ref{fig:mapping_pipeline}.
Head-view gaze annotations are obtained using Aria’s Machine Perception Service \citep{projectaria_tools}, which gives per-frame gaze point in the RGB camera frame. 
To obtain neck-view gaze points, we utilize the VGGT \citep{wang2025vggt} model which is capable of directly inferring cross-view point correspondence by multiple camera frames viewing the same scene, allowing us to map gaze from the head-mounted camera coordinate frame to the neck-mounted camera coordinate frame.
To yield paired frames from head-view and neck-view for each timestep, we synchronized videos from the two cameras using a QR-code–based visual synchronization approach, where at the beginning of each session, participants face a screen displaying frame-unique alternating QR codes visible to both cameras, which later is used to align the frames.
The synchronized head- and neck-view frame pairs and the head-view gaze point is then inputted into VGGT \citep{wang2025vggt}, where we yield the neck-view gaze point using its mapping module.
The model also provides estimates of the relative camera transformation, which we retain as auxiliary annotations.

Each gaze point is further labeled into four categories following standard conventions: fixation, saccade, truncated (out-of-view), and untracked. We apply confidence and visibility thresholds to identify unreliable mappings and classify fixation versus saccade based on inter-frame gaze displacement, following prior work \citep{lai2024glc}. To protect privacy, all faces in the videos are anonymized using EgoBlur \citep{raina2023egoblur}. Finally, all videos are segmented into 5-second clips following \citep{lai2024glc}. In total, we collected approximately 4 hours of effective data, where the detail is shown in Table \ref{tab:task_distribution_time}.

\begin{table}[h]
    \centering
    \caption{Task time distribution of the dataset used in training. Clip chunks are formatted in 5-seconds lengths following \citep{lai2024glc}.}
    \label{tab:task_distribution_time}
    \begin{tabular}{l r r r}
        \toprule
        \textbf{Task} &
        \textbf{Train Time} &
        \textbf{Test Time} &
        \textbf{Total Time} \\
        \midrule
        buy drink            & 3m 40s        & 5m 30s        & 9m 10s        \\
        copy \& cleaning     & 23m 45s       & 13m 15s       & 37m 0s        \\
        eat snack            & 15m 15s       & 2m 35s        & 17m 50s       \\
        make coffee          & 26m 5s        & 12m 10s       & 38m 15s       \\
        prepare food         & 4m 55s        & 6m 15s        & 11m 10s       \\
        jenga               & 1h 27m 15s    & 39m 50s       & 2h 7m 5s      \\
        \midrule
        \textbf{Total}      & 2h 40m 55s    & 1h 19m 35s    & 4h 0m 30s     \\
        \bottomrule
    \end{tabular}
\end{table}

\subsection{Dataset Analysis}
\begin{figure}[h]
    \centering
    \includegraphics[width=0.8\linewidth]{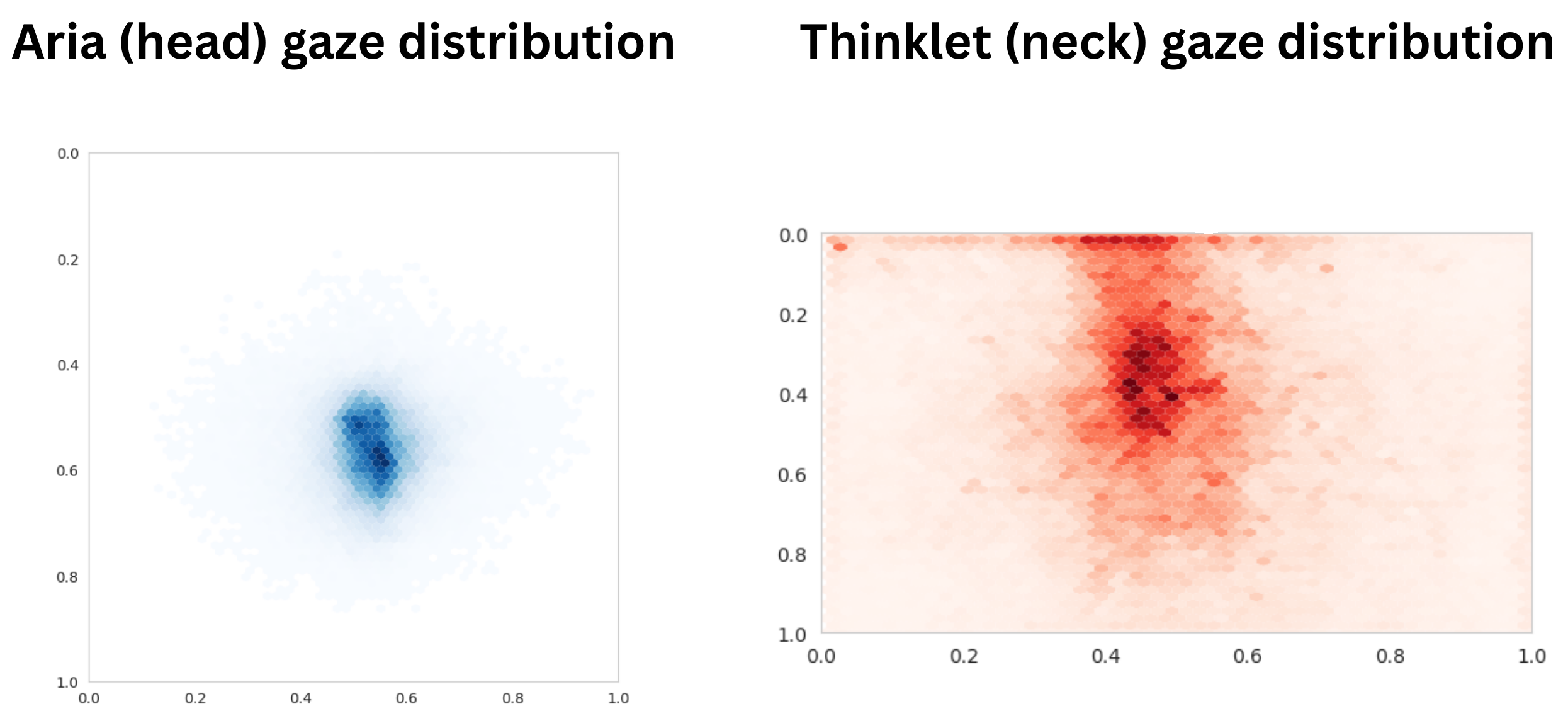}
    \caption{Distribution of the gaze point in head-view camera and neck-view camera.}
    \label{fig:ds_distribution}
\end{figure}

Preliminary analysis on our dataset reveals domain differences between head-mounted and neck-mounted gaze. Notably, 15.9\% of valid neck-mounted gaze points fall outside the camera field of view, compared to 0\% for head-mounted views. Additionally, while head-mounted gaze exhibits strong center bias \citep{li2013learning}, neck-mounted gaze shows a weaker and vertically skewed distribution, with gaze concentrated toward the upper image region (Fig. \ref{fig:ds_distribution}). This can be explained by the neck-mounted camera's optical axis is located below the eye level. 

\section{Experiments}

\subsection{Setup}
\textbf{Backbone.} We use Global--Local Correlation (GLC) \citep{lai2024glc} as the base model due to its strong performance on head-mounted gaze estimation and available pretrained weights on Ego4D \citep{grauman2022ego4d}. 
We compare three variants of the GLC model, namely (i) direct fine-tuning on our dataset, (ii) fine-tuning with an auxiliary in-view classifier, and (iii) multi-view co-learning.

\textbf{Data preprocessing.} Following \citep{lai2024glc}, during training, we sample $T{=}8$ frames at a fixed stride of 8 from each 5-second clip. We apply random spatial augmentation by performing one of the cropping strategy: left crop, center crop or right crop and update gaze coordinates accordingly. During evaluation the temporal sampling is fixed to the start of the clip and spatial augmentation is fixed to center cropping.
Ground-truth heatmaps are generated as a Gaussian centered at the gaze location for in-view frames, and a uniform heatmap for truncated/untracked cases. For the auxiliary in-view classifier, fixation/saccade frames are labeled as in-view and truncated/untracked as out-of-view.

\textbf{Optimization.} All models are trained with AdamW \citep{loshchilov2017adamw} using a learning rate of $10^{-4}$ for 30 epochs, fine-tuning all weights. We use KL-divergence as the heatmap loss $\mathcal{L}_{heatmap}$ (as in \citep{lai2024glc}). 
For the auxiliary classifier approach, we use BCE loss as the in-bound classification loss $\mathcal{L}_{inbound}$. 
For multi-view co-learning approach, we use MSE loss as the alignment loss $\mathcal{L}_{align}$.
In both cases, we use unit weights for the combined objective.

\subsection{Evaluation}
We report adaptive F1 as in \citep{lai2024glc}, selecting the best threshold over a fixed candidate set that maximizes F1 score.
Following the GLC protocol, evaluation is performed on fixation frames with in-view gaze only.

\subsection{Results}
\textbf{Quantitative.} Table~\ref{tab:glc_results} summarizes performance on our dataset. Direct fine-tuning yields an adaptive F1 of 45.2. Adding the auxiliary in-view classifier improves performance to 46.1 (+0.9), while multi-view co-learning underperforms the baseline (44.6). The auxiliary classifier itself achieves 77.8 F1. 

\begin{figure}[h]
    \centering
    \includegraphics[width=0.95\linewidth]{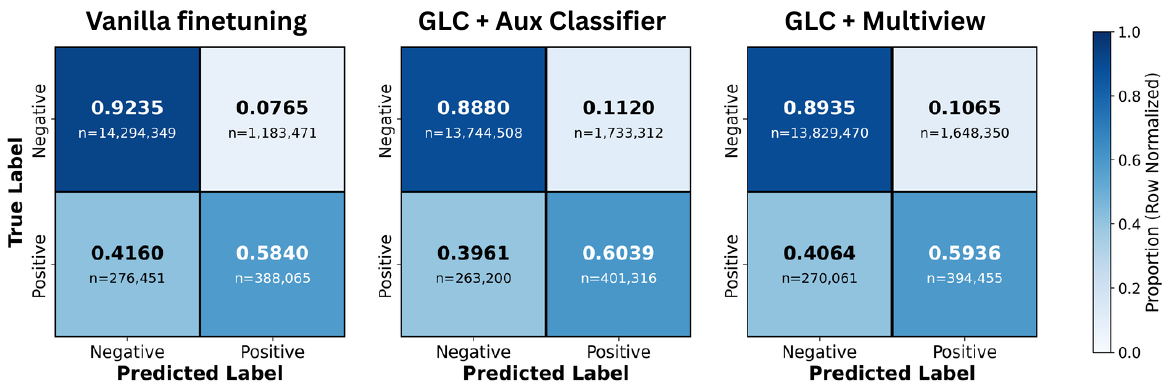}
    \caption{Confusion matrix for the baseline model, auxiliary classifier and multi-view co-learning method.}
    \label{fig:confusion_matrix}
\end{figure}

The confusion matrices for three candidate models are illustrated in Fig. \ref{fig:confusion_matrix}, which suggests that both the auxiliary classifier and the multi-view co-learning method improve model's performance in assigning higher weights to the true gaze region (positive pixels) compared to the baseline, but on contrast loses performance on background pixels (negative pixels).

\begin{table}[ht]
    \centering

    \begin{tabular}{lccc}
        \hline
        \textbf{Model} 
        & \textbf{F1 (\%)} 
        & \textbf{Precision (\%)} 
        & \textbf{Recall (\%)} \\
        \hline

        Vanilla finetuning
        & 45.2
        & 36.7
        & 58.8 \\

        GLC + Aux Classifier
        & \textbf{46.1}
        & \textbf{37.0}
        & \textbf{61.2} \\

        \quad (Classification)
        & (77.8)
        & (93.9)
        & (66.4) \\

        GLC + Multiview
        & 44.6
        & 35.4
        & 60.1 \\
        
        \hline
    \end{tabular}

    \caption{Performance comparison of GLC variants. The auxiliary binary classifier is reported as a sub-component of the GLC + Auxiliary Classifier model. Best results are highlighted in bold.}
    \label{tab:glc_results}
\end{table}

\begin{figure}
    \centering
    \includegraphics[width=0.95\linewidth]{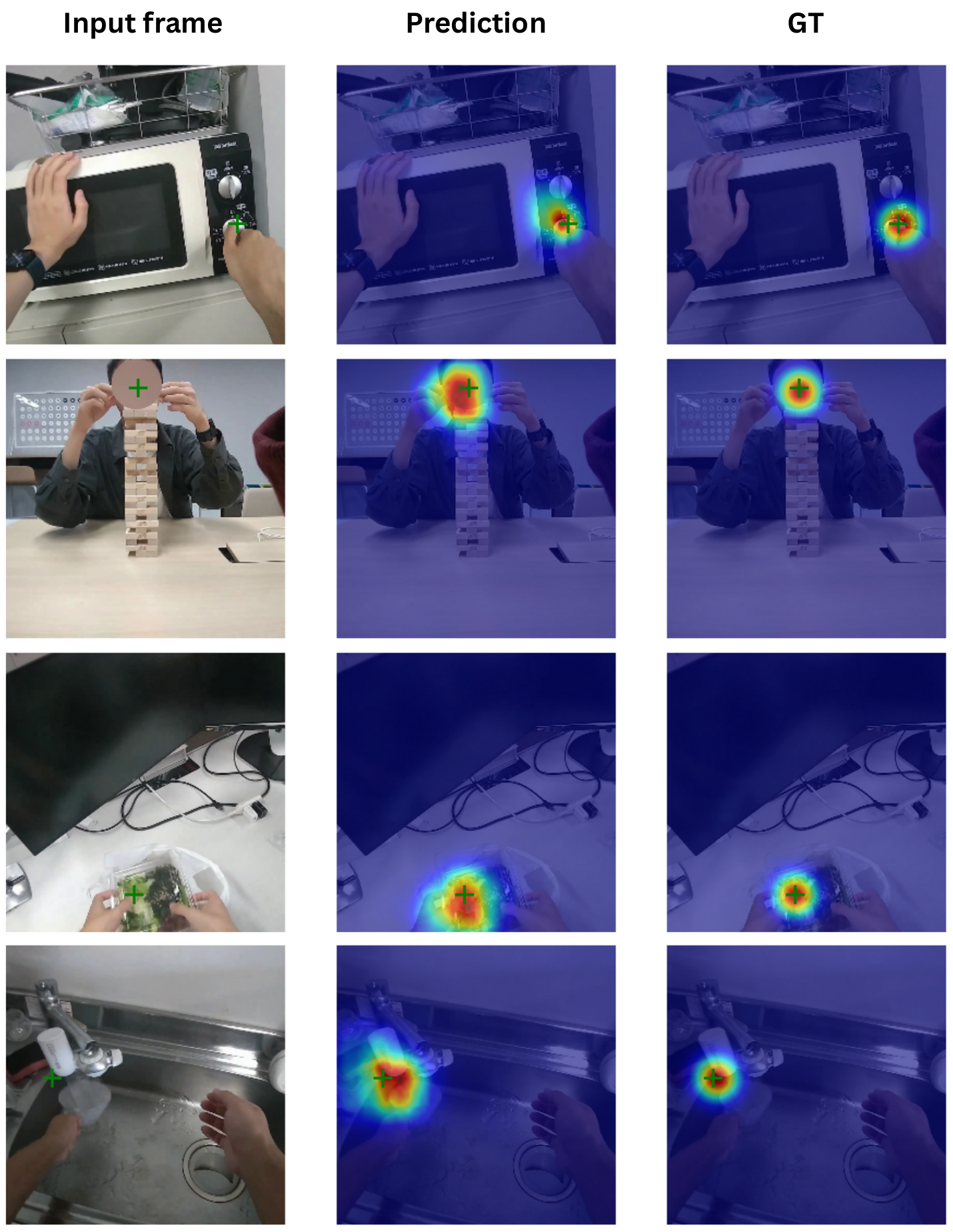}
    \caption{Visualization of the success cases for gaze estimation.}
    \label{fig:success_analysis}
\end{figure}

\begin{figure}
    \centering
    \includegraphics[width=0.95\linewidth]{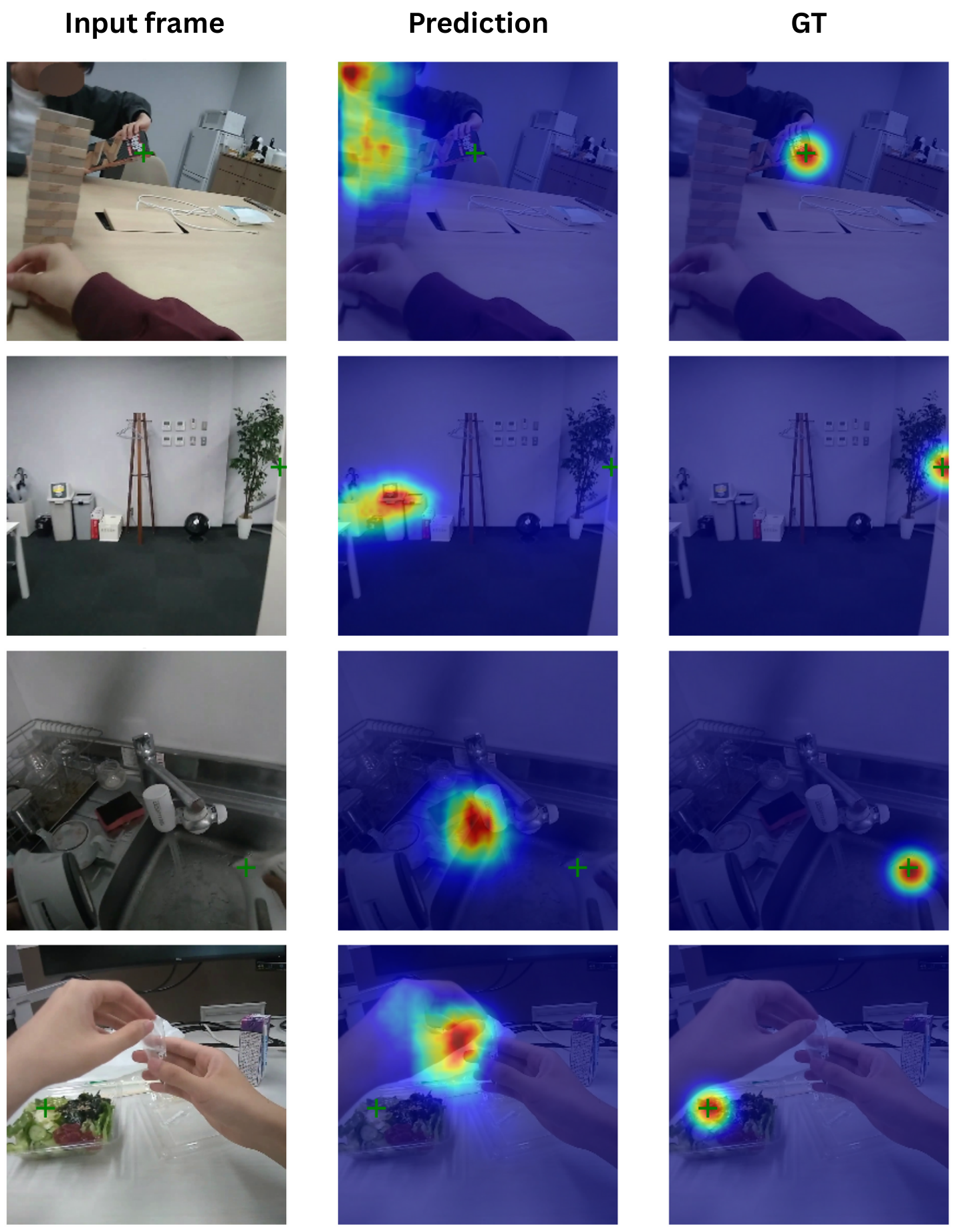}
    \caption{Visualization of the failure cases for gaze estimation.}
    \label{fig:failure_analysis}
\end{figure}

\textbf{Qualitative.} Fig.~\ref{fig:success_analysis}--\ref{fig:failure_analysis} show representative predictions. Successful cases often involve clear hand--object interaction and action context. Failure modes include (i) saliency bias where the model fixates on visually dominant objects despite task context, (ii) hand--gaze mismatch where gaze anticipates future actions rather than current manipulation, and (iii) low-cue frames caused by occlusion or visually uninformative scenes such as empty scenes.

\clearpage

\clearpage

\section{Conclusion and Future Work}

We introduce neck-mounted view gaze estimation, a new egocentric gaze estimation setting that predicts user gaze from a neck-mounted camera perspective. 
Unlike prior work that predominantly relies on head-mounted devices, this task explores gaze estimation under a more comfortable camera placement, enabling new applications in real-world and constrained working environments.

To support this task, we collect and analyze the first dataset for neck-mounted view gaze estimation using a co-collection setup with an eye-tracker–equipped head-mounted camera and a neck-mounted camera. By mapping head-view gaze to the neck-view using VGGT \citep{wang2025vggt}, we obtain gaze annotations and demonstrate domain differences, including frequent out-of-bound gaze and a weaker center bias compared to head-mounted views.

We further study how existing head-mounted gaze models transfer to this new domain. Using GLC \citep{lai2024glc} as a representative baseline, we evaluate two domain-specific augmentations. An auxiliary in-view classification task yields modest improvements, while a multi-view co-learning approach based on latent feature alignment does not improve performance.

Possible future work on this topic includes scaling up data collection to improve generalization, as there is a lack of data comparing to head-mounted view based large scale egocentric datasets. This could be done both by increasing participant diversity and by incorporating a wider range of neck-mounted devices. 
In addition, developing architectures explicitly tailored to neck-mounted views remains an open direction, such as evaluating additional gaze estimation backbones beyond GLC \citep{lai2024glc} or exploring architectures that better adapt to the characteristics of the neck-mounted view domain.

\clearpage
\bibliography{biblio}
\bibliographystyle{colm2024_conference}

\end{document}